\documentclass[11pt]{article}

\usepackage[final]{acl}

\usepackage{times}
\usepackage{latexsym}

\usepackage[T1]{fontenc}

\usepackage[utf8]{inputenc}

\usepackage{microtype}

\usepackage{inconsolata}

\usepackage{graphicx}
\graphicspath{{../}}
\usepackage{booktabs}
\usepackage{multirow}
\usepackage{amsmath,amssymb}
\usepackage{array}
\usepackage{xcolor}
\usepackage{url}
\usepackage{hyperref}
\usepackage{subcaption}
\title{Overview of the PsyDefDetect Shared Task at BioNLP 2026: Detecting Levels of Psychological Defense Mechanisms in Supportive Conversations}

\author{
  Hongbin Na\textsuperscript{1,}\thanks{Co-leads of the shared task organization.} \quad
  Zimu Wang\textsuperscript{2,}\footnotemark[1] \quad
  Zhaoming Chen\textsuperscript{3} \quad
  Yining Hua\textsuperscript{4} \\
  \textbf{Rena Gao}\textsuperscript{5} \quad
  \textbf{Kailai Yang}\textsuperscript{6} \quad
  \textbf{Ling Chen}\textsuperscript{1} \quad
  \textbf{Wei Wang}\textsuperscript{2} \quad
  \textbf{Shaoxiong Ji}\textsuperscript{7,8} \\
  \textbf{John Torous}\textsuperscript{4} \quad
  \textbf{Sophia Ananiadou}\textsuperscript{6} \\[2pt]
 \textsuperscript{1}University of Technology Sydney \quad
  \textsuperscript{2}Xi'an Jiaotong-Liverpool University \\
 \textsuperscript{3}University of Utah \quad
  \textsuperscript{4}Harvard University \quad
  \textsuperscript{5}The University of Melbourne \\
 \textsuperscript{6}The University of Manchester \quad
  \textsuperscript{7}ELLIS Institute Finland \quad
  \textsuperscript{8}University of Turku \\
 \texttt{Hongbin.Na@student.uts.edu.au, Zimu.Wang@liverpool.ac.uk}
}

\begin{document}
\maketitle
\begin{abstract}
We present an overview of \textsc{PsyDefDetect}, the shared task on detecting
levels of psychological defense mechanisms in emotional support dialogues,
co-located with BioNLP@ACL~2026. Grounded in the clinically validated
Defense Mechanism Rating Scales (DMRS) framework,
the task asks systems to classify a target seeker utterance, given its preceding
dialogue context, into one of nine categories: seven hierarchical DMRS levels
plus two auxiliary labels.
Participants worked on \textsc{PsyDefConv}, a
newly released corpus of 200 dialogues and 2{,}336
help-seeker utterances annotated under DMRS with substantial inter-annotator
agreement. The task attracted 172 participants on
CodaBench who produced 563 submissions, with 21 teams
officially registering their results for the final ranking. The best
system achieved a macro F1-score of 0.420,
surpassing the strongest fine-tuned baseline reported in the dataset paper by a notable margin, yet leaving clear headroom.
Our analysis highlights (i) a persistent tendency to over-predict the majority
\emph{High-Adaptive} class, (ii) a widening gap between accuracy and macro-F1
that reveals class-imbalance sensitivity, and (iii) the value of theory-aware
and LLM-based approaches for fine-grained defensive-function classification.
We release all task materials and invite the community to continue work on this
novel intersection of clinical psychology and NLP.\footnote{Task website:
\url{https://psydefdetect-shared-task.github.io/}; CodaBench:
\url{https://www.codabench.org/competitions/12124/}.}
\end{abstract}

\section{Introduction}
\label{sec:intro}
 
Psychological defenses are, in Winnicott's words, the means by which the
\emph{False Self}, if successful in its function, hides the \emph{True Self}
\cite{winnicott2018ego}. They are automatic strategies people use to regulate
distress, and when used rigidly or excessively they are linked to poorer mental
health outcomes and to interpersonal difficulties \cite{perry2004studying, di2024therapists}. In emotional support conversations (ESC) \cite{liu-etal-2021-towards}, defenses shape what help-seekers disclose and how they accept or resist help---yet, despite rapid progress on empathy modeling \cite{hua2025charting, cai-etal-2024-empcrl,info:doi/10.2196/52597}, strategy selection \cite{kang-etal-2024-large,hua2025scoping,na-etal-2025-survey}, and affect understanding \cite{wang-etal-2024-knowledge,wang-etal-2025-posts,ma-etal-2025-detecting,zhao2025hears}, the defensive function of seeker utterances
remains largely unmodeled in current ESC systems \cite{di2024therapists}.
 
To catalyze research on this under-explored dimension of supportive dialogue,
we organized \textbf{\textsc{PsyDefDetect}}, a shared task co-located with
BioNLP@ACL~2026 on detecting \emph{levels} of psychological defense
mechanisms. The task is grounded in the clinically validated Defense
Mechanism Rating Scales (DMRS) \cite{perry2004studying,vaillant2012adaptation}, and asks systems to classify each help-seeker utterance---given its preceding
dialogue context---into one of nine categories: the seven hierarchical DMRS
levels, augmented with two auxiliary labels for phatic and under-specified turns. The task is built on \textsc{PsyDefConv} \cite{na-etal-2026-psydefconv}, the first conversational corpus annotated with DMRS-based defense levels.
\textsc{PsyDefConv} is derived from ESConv \cite{liu-etal-2021-towards} via stratified sampling over problem types and emotions, and was double-blind annotated to substantial agreement.
 
\paragraph{Participation.}
\textsc{PsyDefDetect} attracted strong community engagement. The task was
hosted on CodaBench \cite{codabench} from December~2025 through April~2026. Over the course of the evaluation period, the competition registered \textbf{172 participants} who together produced \textbf{563 submissions} to the leaderboard. By the final registration deadline, \textbf{21 teams} had officially submitted their results for inclusion in the ranking. The best system reached a macro F1-score of 0.420, substantially surpassing the strongest fine-tuned baseline
reported in the dataset paper ($\approx 0.315$ macro-F1 \cite{na-etal-2026-psydefconv}), while still leaving ample headroom for future research.
 
\paragraph{Contributions.}
This paper reports on the organization, dataset, participating systems, and
results of \textsc{PsyDefDetect}. Our contributions are as follows:
\begin{itemize}
    \item We introduce \textsc{PsyDefDetect}, the first shared task that
    operationalizes DMRS-based defensive functioning as an utterance-level
    classification problem over emotional support dialogue.
    \item We report task-level participation that reflects substantial
    community engagement, with 172 CodaBench participants, 563 leaderboard
    submissions, and 21 officially ranked teams.
    \item We present a methodological taxonomy of participating systems
    together with a full leaderboard analysis that benchmarks submissions
    against both zero-shot and fine-tuned baselines from the dataset paper.
    \item We characterize the dominant failure modes observed across the
    leaderboard, in particular the severe sensitivity to class imbalance and
    the systematic over-prediction of the High-Adaptive level, and use them
    to motivate concrete directions for future work at the intersection of
    clinical psychology and NLP.
\end{itemize}
 
\section{Background and Related Work}
\label{sec:related}
 
\paragraph{Defense mechanisms and DMRS.}
Defense mechanisms are a cornerstone of psychodynamic theory
\cite{freud1936inhibitions}. The \emph{Defense Mechanism Rating Scales} (DMRS)
organize roughly thirty mechanisms into seven hierarchical levels of
defensive maturity, from \emph{Action Defenses} (Level~1) to
\emph{High-Adaptive Defenses} (Level~7) \cite{perry2004studying,vaillant2012adaptation}.
DMRS was originally designed for longitudinal clinical case formulation rather
than for single-utterance classification; \textsc{PsyDefConv}
\cite{na-etal-2026-psydefconv} adapts the scheme to conversational text by
annotating at the \emph{level} rather than the \emph{mechanism} granularity,
increasing identifiability and reliability.
 
\paragraph{Emotional support conversations.}
Research on ESC has built increasingly capable dialogue agents that alleviate
user distress through multi-turn, strategy-grounded interaction, starting
with the ESConv corpus \cite{liu-etal-2021-towards} and extending to multi-strategy
\cite{bai2025emotionalsupportersusemultiple}, synthetic \cite{zheng-etal-2023-augesc,wang-etal-2024-knowledge}, and reasoning-aware \cite{zhang-etal-2024-escot} variants.
These efforts have largely focused on \emph{supporter} strategies, empathy,
and affect modeling \cite{wang-etal-2024-knowledge,cai-etal-2024-empcrl,info:doi/10.2196/52597,hua2025charting,xu2026harm}, while
the \emph{defensive function} of seeker utterances has remained unmodeled.
 
\paragraph{Mental health classification.}
Prior shared tasks at BioNLP and CLPsych have addressed clinically-motivated
classification problems such as suicide risk assessment, depression detection,
and empathy prediction.
BioNLP has covered a wide spectrum of medical-related challenges, such as clinical question-answering and summarization \cite{colelough-etal-2025-overview,soni-etal-2025-overview,xiao-etal-2025-overview}. However, tasks specifically addressing mental health have remained notably underrepresented.
CLPsych shared tasks have addressed a range of mental-health-related problems, including depression and PTSD \cite{coppersmith-etal-2015-clpsych}, suicide risk assessment \cite{zirikly-etal-2019-clpsych,chim-etal-2024-overview}, online peer support \cite{milne-etal-2016-clpsych}, developmental mental-health prediction \cite{lynn-etal-2018-clpsych}, and longitudinal affective modeling \cite{tsakalidis-etal-2022-overview,tseriotou-etal-2025-overview}. However, these efforts have primarily centered on social media platforms such as X and Reddit, with comparatively limited attention to ESC.
\textsc{PsyDefDetect}
complements this line of work by introducing a theory-grounded, fine-grained
classification problem that requires both local linguistic reasoning and
context-aware interpretation.
 
\section{Task Description}
\label{sec:task}
 
\subsection{Problem Formulation}
Given a multi-turn emotional support dialogue
$D = (u_1, u_2, \dots, u_t)$ between a help-\emph{seeker} and a
\emph{supporter}, and a target help-seeker utterance $u_t$, the task is to
predict a defense level label $y \in \mathcal{Y}$, where $\mathcal{Y}$
comprises the seven DMRS levels and two auxiliary labels:
 
\begin{itemize}
    \item \textbf{Level 0 -- No Defenses:} phatic or functional utterances
          that do not engage with psychological conflict.
    \item \textbf{Level 1 -- Action Defenses:} passive aggression, help-rejecting
          complaining, acting out.
    \item \textbf{Level 2 -- Major Image-Distorting:} splitting, projective identification.
    \item \textbf{Level 3 -- Disavowal:} denial, rationalization, projection, autistic fantasy.
    \item \textbf{Level 4 -- Minor Image-Distorting:} devaluation, idealization, omnipotence.
    \item \textbf{Level 5 -- Neurotic:} repression, dissociation, reaction formation, displacement.
    \item \textbf{Level 6 -- Obsessional:} isolation of affect, intellectualization, undoing.
    \item \textbf{Level 7 -- High-Adaptive:} affiliation, altruism, anticipation,
          humor, self-assertion, self-observation, sublimation, suppression.
    \item \textbf{Level 8 -- Needs More Information:} context is insufficient to
          assign a label with confidence.
\end{itemize}
 
Systems have access to the dialogue context preceding and including $u_t$,
but must not use future turns---this mirrors the online nature of the
clinical annotation setup \cite{na-etal-2026-psydefconv}.
 
 
 
\section{The \textsc{PsyDefConv} Dataset}
\label{sec:data}
 
\subsection{Construction}
\textsc{PsyDefConv} \cite{na-etal-2026-psydefconv} is built on top of ESConv
\cite{liu-etal-2021-towards}. The organizers performed stratified sampling over
the joint distribution of problem types and emotions in ESConv to obtain a
representative 200-dialogue subset, annotated at the \emph{seeker}-utterance
level for defense functioning.
 
Two trained annotators with expertise in both psychology and NLP labeled
each seeker turn independently and double-blind, reaching
Cohen's $\kappa = 0.639$ (substantial agreement). Disagreements were
adjudicated by consensus to form the gold standard. The annotation workflow
was supported by \textsc{DMRS Co-Pilot} \cite{na-etal-2026-psydefconv}, a
four-stage LLM pipeline that produces stressor hypotheses, screens candidate
DMRS items, validates evidence, and synthesizes ranked recommendations; it
reduced mean annotation time by 24.0\%.
 
\subsection{Corpus Statistics}
\label{sec:data:stats}
 
Table~\ref{tab:basic_stats} summarizes the basic statistics of
\textsc{PsyDefConv}. The corpus contains 200 dialogues and 4{,}709
utterances, roughly evenly divided between supporter (2{,}373) and seeker
(2{,}336) turns. Conversations are of moderate length, with a mean of
23.5 turns per dialogue and an average utterance length of 19.8 tokens.
Only the 2{,}336 seeker turns carry DMRS defense-level annotations;
supporter turns are provided as dialogue context but are not part of the
evaluation.
 
\begin{table}[t]
\centering
\resizebox{\columnwidth}{!}{%
\begin{tabular}{lrrr}
\toprule
\textbf{Category} & \textbf{Total} & \textbf{Supporter} & \textbf{Seeker} \\
\midrule
\# Dialogues & 200 & -- & -- \\
\# Utterances & 4,709 & 2,373 & 2,336 \\
Avg. Turns per Dialogue & 23.5 $\pm$ 6.6 & 11.9 $\pm$ 3.4 & 11.7 $\pm$ 3.3 \\
Avg. Length of Utterances & 19.8 $\pm$ 16.5 & 20.9 $\pm$ 17.0 & 18.8 $\pm$ 15.8 \\
\bottomrule
\end{tabular}%
}
\caption{Data statistics of \textsc{PsyDefConv}.}
\label{tab:basic_stats}
\vspace{-2mm}
\end{table}
 
\subsection{Label Distribution}
\label{sec:data:dist}
 
Table~\ref{tab:psydef_distributions} reports the distribution of the
2{,}336 annotated seeker utterances across the nine defense labels, as well
as their aggregation into four broader defensive categories. The
distribution is strongly skewed toward the \emph{High-Adaptive} level
(Level~7), which alone accounts for 51.8\% of all annotated utterances.
Non-defensive or contextually ambiguous turns (Levels~0 and 8) jointly
cover another 17.4\%, leaving only roughly one third of the corpus for
the remaining six defense levels. Within that tail, \emph{Immature}
defenses (Levels~1--4) aggregate to 18.9\% while \emph{Neurotic} defenses
(Levels~5--6) account for 11.9\%, with the individual minority levels
ranging from 2.6\% (Neurotic) to 9.2\% (Obsessional).
 
This imbalance reflects the natural prevalence of defensive functioning in
supportive dialogue \cite{na-etal-2026-psydefconv}: adaptive coping such as
self-assertion, self-observation, and seeking affiliation is, by design,
what participants in ESC conversations most often express. However, the
imbalance also means that systems optimized for overall accuracy can
trivially favor the dominant class at the expense of minority defense
levels. We adopt macro-averaged F1 over the positive classes (1--8) as the
official ranking metric (Section~\ref{sec:eval}) precisely to penalize such
majority-class bias and to encourage systems that discriminate well across
the full DMRS hierarchy.
 
\begin{table}[t]
  \centering
  \footnotesize
  \resizebox{\columnwidth}{!}{%
    \begin{tabular}{lrr}
      \toprule
      \textbf{Categories} & \textbf{Num} & \textbf{Proportion} \\
      \midrule
      \multicolumn{3}{c}{\textbf{Seeker's Defense Levels}} \\
      \midrule
      0 No Defenses                     & 371   & 15.9\% \\
      1 Action Defenses                 & 136   & 5.8\% \\
      2 Major Image-Distorting          & 77    & 3.3\% \\
      3 Disavowal                       & 124   & 5.3\% \\
      4 Minor Image-Distorting          & 105   & 4.5\% \\
      5 Neurotic                        & 61    & 2.6\% \\
      6 Obsessional                     & 216   & 9.2\% \\
      7 High-Adaptive                   & 1{,}211 & 51.8\% \\
      8 Needs More Information          & 35    & 1.5\% \\
      \midrule
      Overall                           & 2{,}336 & 100.0\% \\
      \midrule
      \multicolumn{3}{c}{\textbf{Seeker's Defense Categories}} \\
      \midrule
      Non-Defensive/Ambiguous (0, 8)    & 406   & 17.4\% \\
      Mature Defenses (7)               & 1{,}211 & 51.8\% \\
      Neurotic Defenses (5, 6)          & 277   & 11.9\% \\
      Immature Defenses (1, 2, 3, 4)    & 442   & 18.9\% \\
      \midrule
      Overall                           & 2{,}336 & 100.0\% \\
      \bottomrule
    \end{tabular}%
  }
  \caption{Distribution of the \textsc{PsyDefConv} dataset across
  annotated defense levels and aggregated defense categories.}
  \label{tab:psydef_distributions}
  \vspace{-2mm}
\end{table}

\section{Evaluation Setup}
\label{sec:eval}
 
\subsection{Platform and Protocol}
The shared task was hosted on CodaBench~\cite{codabench}, where
participants downloaded the dataset, received starter baseline kits, and
submitted predictions as zipped JSON files. Each team could register on the
official leaderboard after completing the result-registration form.
 
\subsection{Metrics}
Following the dataset paper \cite{na-etal-2026-psydefconv}, we report
Accuracy, Macro Precision, Macro Recall, and
Macro F1-score. Macro F1 evaluated over the positive
classes (1--8) is the official ranking metric, chosen because it penalizes
majority-class bias, weighs minority defensive levels equally with the
dominant High-Adaptive level, and follows established practices for multi-class classification \cite{wang-etal-2022-maven,wang-etal-2024-document-level,chen-etal-2025-medfact}.
Level~0 is excluded from the official macro-F1 because it marks phatic or
functional turns without defensive content. Level~8, by contrast, is retained
because detecting that the available context is insufficient is an explicit
task decision rather than a negative class.
Furthermore, we include an additional leaderboard that aggregates performance across all classes, enabling a more comprehensive and holistic comparison of the submitted systems.
 
\subsection{Baselines}
The CodaBench page provided zero-shot prompting baselines using strong
general-purpose LLMs. These baselines follow the same setup as the dataset
paper; for reference, the strongest results reported in
\cite{na-etal-2026-psydefconv} are as follows. Under zero-shot
prompting, Gemini 2.5 Pro reaches an accuracy of 0.5636 and a macro-F1 of
0.2599, while DeepSeek-V3.2 with thinking enabled achieves comparable
performance at 0.5572 accuracy and 0.2617 F1. Under supervised fine-tuning,
the best results come from Ministral-8B (Acc~=~0.6483, F1~=~0.3148) and
InternLM3-8B (Acc~=~0.6398, F1~=~0.3053).

 
\section{Participating Systems}
\label{sec:systems}
 
\subsection{Participation Statistics}
The \textsc{PsyDefDetect} shared task drew broad international
engagement. The CodaBench competition registered \textbf{172
participants} and received \textbf{563 submissions} over the course of
the evaluation period, with \textbf{21 teams} officially registering
their results for the final ranking
(Table~\ref{tab:leaderboard}).
Of these, \textbf{15} accompanied their submission with a system
description paper contributed to this volume. The author affiliations on
these 15 papers span 12 countries---Australia, Bangladesh, Canada,
China, Germany, India, Russia, Spain, Switzerland, the UK, the USA,
and Vietnam---and 24 distinct institutions, including a number of
cross-institution collaborations that combine academic NLP groups
with clinical and biomedical informatics partners.
 
\subsection{Methodological Taxonomy}
\label{sec:systems:taxonomy}
The 15 described systems span a broad methodological space, but
gravitate toward six recurring patterns, described below. Almost every
team targets the same two task-specific pain points: the heavy skew
toward Level~7 (``High-Adaptive''), and the ambiguity of DMRS boundaries
for short, context-dependent utterances. Several teams combine two or
more strategies, in which case we record the most distinctive
contribution here.
 
\paragraph{(i) Multi-model ensembles and deliberative-agent architectures.}
Top-ranked systems reject the idea that a stronger single model can solve
the task, and instead explicitly engineer \emph{error independence}
across voters or agents. \textsc{Nürnberg NLP} (ranked 1st;
\citealp{steigerwald-etal-2026-nuernbergnlp}) constructs a
9-voter ensemble along three orthogonal axes---class granularity
(9-class \emph{gatekeeper} vs.\ 8-class \emph{specialists}), training
paradigm (generative vs.\ discriminative), and base model (Ministral
vs.\ Phi-4)---with per-axis cross-validation folds. \textsc{UTS}
(2nd; \citealp{galat-rizoiu-2026-uts}) operates a multi-phase deliberative
council of Gemini 2.5 agents in which class-specific \emph{advocates}
rate evidence strength rather than vote, augmented by a targeted
override ensemble of three fine-tuned Qwen-family models. \textsc{TONI-NLP}
(9th; \citealp{paul-etal-2026-toninlp}) likewise finds that ensembles
outperform any individual prompting, fine-tuning, or
embedding-classifier baseline.
 
\paragraph{(ii) Retrieval- and rubric-grounded LLM classification.}
A second cluster of systems brings the DMRS clinical rubric into the
prompt, either dynamically via retrieval or statically via curated
boundary cues. \textsc{PerceptionLab}
(3rd; \citealp{fahim-etal-2026-perceptionlab}) pairs dynamic DMRS-Q item
retrieval (Gemini 2.5 Pro) with a Gemini 2.5 Flash classifier fine-tuned
on reasoning traces distilled from the same Pro model, explicitly
targeting ``LLM polarization'' toward extreme labels.
\textsc{DAL~Team} (10th; \citealp{tran-etal-2026-dalteam}) runs a
retrieval-augmented LLM pipeline that decomposes prediction into a
coarse-to-fine hierarchy and adds summary-based distillation of
dialogue context.
\textsc{zzucs} (6th; \citealp{huang-etal-2026-zzucs}) uses its
\emph{CoR-QLoRA} approach to encode task contracts, taxonomy definitions,
and adjacent-level boundary cues into the prompts used for 8B QLoRA
adaptation.
 
\paragraph{(iii) Parameter-efficient LLM fine-tuning.}
Most mid-to-top-ranked systems converge on QLoRA or equivalent PEFT on
mid-scale open LLMs. \textsc{LinguIUTics}
(4th; \citealp{adib-etal-2026-linguiutics}) QLoRA-fine-tunes
Qwen3-8B with grouped stratified cross-validation, minority-class
lexical augmentation, and a post-hoc logit-bias calibration step that
substantially improves the rare ``Needs More Information'' class. \textsc{Eraserhead} (7th; \citealp{horaira-etal-2026-eraserhead})
fine-tunes Qwen3-14B with clinically informed prompts and iteratively
retuned per-class oversampling. \textsc{transformer\_1376}
(12th; \citealp{saha-etal-2026-transformer1376}) reformulates the task
as conditional text generation and applies 4-bit QLoRA to Gemma-2-2B.
 
\paragraph{(iv) Encoder-only and domain-specific transformer fine-tuning.}
Three teams explore encoder backbones. \textsc{Neural Nexus}
(11th; \citealp{basu-2026-neuralnexus}) fine-tunes \textsc{roberta-base}
with a composite objective combining focal loss, label smoothing, and
square-root-dampened class weights, using role-tagged dialogue history
and a \texttt{[TARGET]} marker on the target utterance.
\textsc{CS\_Metro} (15th; \citealp{rebayet-etal-2026-csmetro}) builds a
three-stage pipeline of LLM-based dialogue summarization, domain-specific
transformer fine-tuning (including Mental-BERT and Mental-RoBERTa variants),
and rule-based ensembling; they report that domain-specific encoders
outperform generic LLM fine-tuning on this clinical task.
\textsc{KCL-Cogstack} (17th; \citealp{agarwal-etal-2026-kclcogstack})
contrasts flat fine-tuning, few-shot prompting, and a hierarchical
coarse-to-fine classifier that exploits the DMRS label tree, finding
the hierarchical design the most effective of the three.
 
\paragraph{(v) Synthetic data and theory-informed augmentation.}
\textsc{vishc} (13th; \citealp{vu-pham-2026-vishc}) focuses on data
scarcity rather than model scale, proposing stressor-anchored,
theory-driven synthetic data generation combined with a hybrid model
that fuses language representations with structured clinical features.
Lexical augmentation and oversampling are also used as secondary
ingredients by \textsc{LinguIUTics}, \textsc{zzucs},
\textsc{transformer\_1376}, and \textsc{Eraserhead}.
 
\paragraph{(vi) Systematic exploration and negative-result studies.}
Two teams emphasize breadth of comparison over a single headline
system. \textsc{AlienAnnotators}
(19th; \citealp{karip-hossain-2026-alienannotators}) systematically
evaluates six open-source small language models ($\leq 9$B) under
zero-shot and fine-tuning regimes, finding that clinically grounded
prompts consistently outperform bare label definitions and that
model scale alone does not help zero-shot performance; their
post-submission configuration (fine-tuning with 5-fold CV and logit
averaging) reaches macro-F1~=~0.346, more than double their official
submission. \textsc{explainators}
(20th; \citealp{babakova-etal-2026-explainators}) similarly runs four
exploratory tracks---direct prompting, encoder fine-tuning, novel
``state-of-mind'' generation, and LLM fine-tuning---across
DeepSeek-V3.2, Qwen-family models, and GLM-series models.

\section{Results}
\label{sec:results}
 
\subsection{Official Leaderboard}
Table~\ref{tab:leaderboard} lists the final leaderboard, ranked by macro
F1-score on the held-out test set. The top system (Nürnberg NLP, F1 = 0.4200)
outperforms the second-ranked team (UTS, F1 = 0.4055) by a small margin, while
the overall spread of F1 scores is wide (0.063--0.420), underscoring the
difficulty of the task.

\begin{table*}[t]
\centering
\small
\setlength{\tabcolsep}{4pt}
\renewcommand{\arraystretch}{1.05}
\begin{tabular}{r l c c c c c c c}
\toprule
 & & & \multicolumn{3}{c}{\textbf{Positive classes (1--8)}}
     & \multicolumn{3}{c}{\textbf{All classes (0--8)}} \\
\cmidrule(lr){4-6}\cmidrule(lr){7-9}
\textbf{Rank} & \textbf{Team} & \textbf{Acc}
   & \textbf{P} & \textbf{R} & \textbf{F1}
   & \textbf{P} & \textbf{R} & \textbf{F1} \\
\midrule
1  & N\"urnberg NLP        & 0.7013 & 0.4510 & 0.4036 & 0.4200 & 0.4959 & 0.4639 & \textbf{0.4732} \\
2  & UTS                   & 0.6737 & 0.4607 & 0.3884 & 0.4055 & 0.4915 & 0.4327 & 0.4450 \\
3  & PerceptionLab         & 0.6737 & 0.4263 & 0.4089 & 0.3956 & 0.4721 & 0.4479 & 0.4402 \\
4  & LinguIUTics           & 0.6419 & 0.4003 & 0.3958 & 0.3917 & 0.4416 & 0.4570 & 0.4427 \\
5  & LDI Lab               & 0.6356 & 0.3770 & 0.3892 & 0.3713 & 0.4277 & 0.4422 & 0.4244 \\
6  & zzucs                 & 0.6441 & 0.3969 & 0.3520 & 0.3585 & 0.4402 & 0.4166 & 0.4135 \\
7  & Eraserhead            & 0.6462 & 0.4075 & 0.3193 & 0.3418 & 0.4482 & 0.3801 & 0.3947 \\
8  & zzunlp                & 0.6758 & 0.4991 & 0.2891 & 0.3300 & 0.5381 & 0.3578 & 0.3909 \\
9  & TONI-NLP              & 0.6716 & 0.4701 & 0.2839 & 0.3196 & 0.5094 & 0.3560 & 0.3813 \\
10 & DAL team              & 0.4831 & 0.4187 & 0.2773 & 0.3113 & 0.4165 & 0.3516 & 0.3391 \\
11 & Neural Nexus          & 0.5169 & 0.2480 & 0.2867 & 0.2556 & 0.3140 & 0.3259 & 0.3080 \\
12 & transformer\_1376     & 0.5508 & 0.2669 & 0.2351 & 0.2475 & 0.3132 & 0.2890 & 0.2979 \\
13 & VISHC                 & 0.5826 & 0.2588 & 0.2503 & 0.2462 & 0.3168 & 0.3069 & 0.3045 \\
14 & GMU-AIT626            & 0.5805 & 0.2844 & 0.2328 & 0.2455 & 0.3298 & 0.2839 & 0.2952 \\
15 & CS\_Metro             & 0.6377 & 0.3001 & 0.2572 & 0.2346 & 0.3673 & 0.3131 & 0.3004 \\
16 & Uprm                  & 0.5169 & 0.2600 & 0.2161 & 0.2288 & 0.3081 & 0.2854 & 0.2877 \\
17 & KCL-Cogstack          & 0.4322 & 0.2913 & 0.2517 & 0.2278 & 0.3359 & 0.3171 & 0.2868 \\
18 & sp\_001               & 0.6398 & 0.3057 & 0.2309 & 0.2253 & 0.3617 & 0.3060 & 0.2953 \\
19 & AlienAnnotators       & 0.5996 & 0.1555 & 0.1975 & 0.1628 & 0.2027 & 0.2823 & 0.2251 \\
20 & explainators          & 0.6144 & 0.2366 & 0.1660 & 0.1612 & 0.3041 & 0.2275 & 0.2296 \\
21 & Gradient Descender    & 0.2606 & 0.1383 & 0.0455 & 0.0629 & 0.1465 & 0.1501 & 0.0946 \\
\midrule
\multicolumn{2}{r}{\emph{Baseline} -- Ministral-8B (fine-tuned) \cite{na-etal-2026-psydefconv}}
  & 0.6483 & 0.3397 & 0.3045 & 0.3148 & 0.3978 & 0.3640 & 0.3745 \\
\multicolumn{2}{r}{\emph{Baseline} -- Gemini 2.5 Pro (zero-shot) \cite{na-etal-2026-psydefconv}}
  & 0.5636 & 0.2749 & 0.2612 & 0.2599 & 0.2907 & 0.3107 & 0.2893 \\
\bottomrule
\end{tabular}
\caption{Official \textsc{PsyDefDetect} leaderboard. Systems are ranked
by macro-F1 over the \emph{positive} classes 1--8 (the official metric
of the shared task); for reference we additionally report
macro-averaged precision, recall, and F1 computed over \emph{all} nine
classes (0--8). Accuracy is shared across both metric groups. The
winning system's official macro-F1 of 0.4200 exceeds the strongest
fine-tuned baseline reported in the dataset paper by ${\approx}10.5$
absolute points; its all-class F1 of 0.4732 exceeds that baseline by
${\approx}9.9$ points.}
\label{tab:leaderboard}
\end{table*}

\subsection{Key Observations}
 

\paragraph{Top systems substantially outperform established baselines.}
The best-performing system, introduced by N\"urnberg NLP, achieves a macro F1-score of 0.420, surpassing the best fine-tuned baseline reported in the dataset paper (Ministral-8B, F1 = 0.315) by approximately \textbf{10.5 absolute points}, and the best zero-shot baseline by \textbf{16 points}. These gains suggest that participants were able to leverage additional task-specific signals beyond standard fine-tuning setups. Nevertheless, the overall performance ceiling remains modest in absolute terms, indicating that the task continues to pose significant challenges.

\paragraph{Wide variability in system performance.}
System performance exhibits a wide dispersion, with macro F1-scores ranging from 0.063 (\textsc{Gradient Descender}) to 0.420 (\textsc{Nürnberg NLP}), corresponding to a $6.7\times$ difference. The distribution further reveals a clear quartile structure: the top quartile achieves F1-scores of $\geq 0.37$, the median lies $\approx 0.25$, and the bottom quartile falls $\leq 0.23$. This pronounced spread indicates that, despite the intrinsic difficulty of the task, there remains considerable room for methodological differentiation and performance gains across approaches.

\begin{figure}[t]
    \centering
    \includegraphics[width=\linewidth]{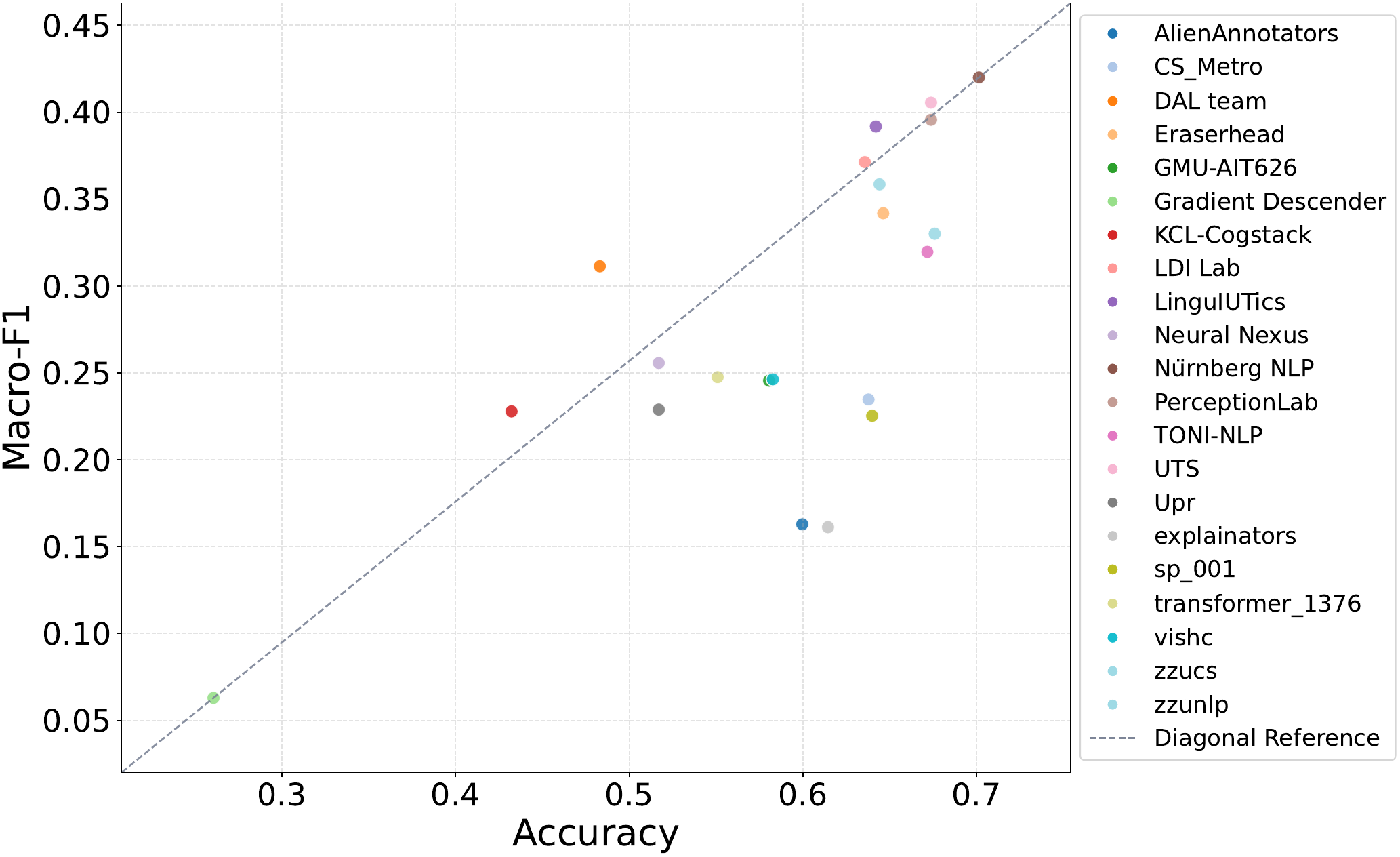}
    \caption{Scatter plot illustrating the relationship between accuracy and macro F1-scores across all submitted systems.}
    \label{fig:scatter}
\end{figure}

\paragraph{Accuracy is a misleading proxy for F1-score.}
Several submissions attain relatively high accuracy while exhibiting substantially lower F1-scores, as shown in Figure~\ref{fig:scatter}. For instance, \textsc{zzunlp} (Acc = 0.676, F1 = 0.330) and \textsc{TONI-NLP} (Acc = 0.672, F1 = 0.320) demonstrate this discrepancy. This mismatch stems from the pronounced class imbalance, with the Level~7 majority class accounting for $\approx$52\% of the dataset. Models that over-predict the High-Adaptive class can achieve competitive accuracy, but suffer from poor recall on minority classes, thereby degrading macro F1-scores. In contrast, systems such as \textsc{Nürnberg NLP}, \textsc{PerceptionLab}, and \textsc{LinguIUTics} achieve a more favorable balance, maintaining strong accuracy alongside improved recall across all classes.
 
 
 
 
\section{Analysis and Discussion}
\label{sec:analysis}
 
\subsection{Per-Class Performance}
 
\begin{figure}[t]
    \centering
    \includegraphics[width=\linewidth]{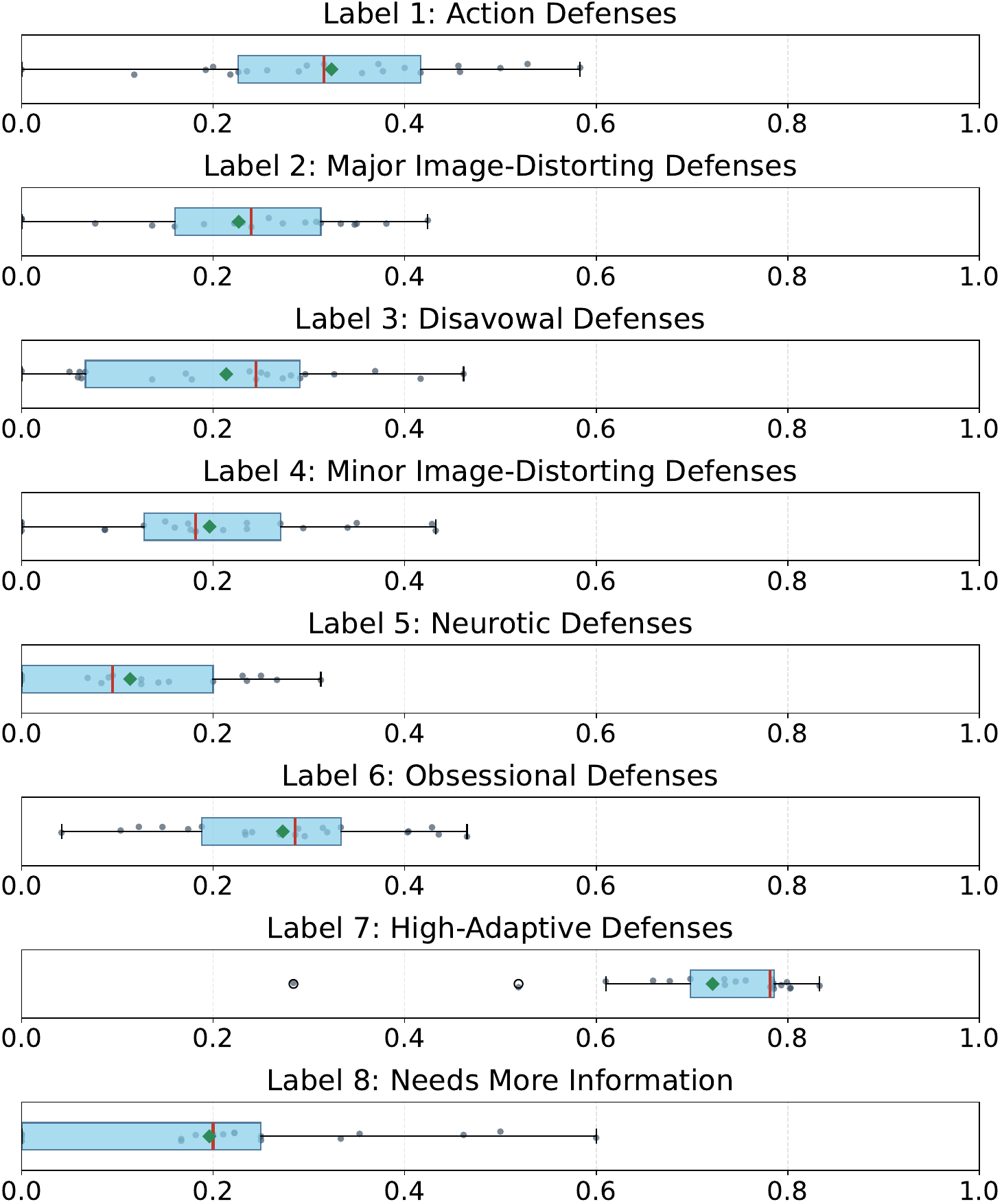}
    \caption{Per-class F1-scores across all submitted systems for the positive
    classes 1--8. Each
    box summarizes, for a given defense level, the distribution of F1
    values over the 21 leaderboard submissions.}
    \label{fig:boxplots}
\end{figure}
 
Figure~\ref{fig:boxplots} summarizes per-class F1 across the 21
submitted systems. Level~7 (High-Adaptive) stands apart: its median
F1 is by far the highest and the inter-team spread is narrow,
reflecting both its majority status (${\approx}52\%$ of the corpus)
and the distinctiveness of its surface markers. Every other class has
a low median and a wide spread---widest for Levels~2--4, whose
linguistic realizations overlap in affective-blame and
reality-distortion cues, and for Level~8 (Needs More Information), on
which a non-trivial fraction of systems collapse to zero. The narrow
box for Level~7 marks a class where surface cues suffice; the
wide boxes (Levels~1--6) mark classes where methodological choices
translate directly into outcome differences.
 
\subsection{Error Patterns}
 
\begin{figure*}[t!]
    \centering
    \begin{subfigure}[b]{0.49\linewidth}
        \includegraphics[width=\textwidth]{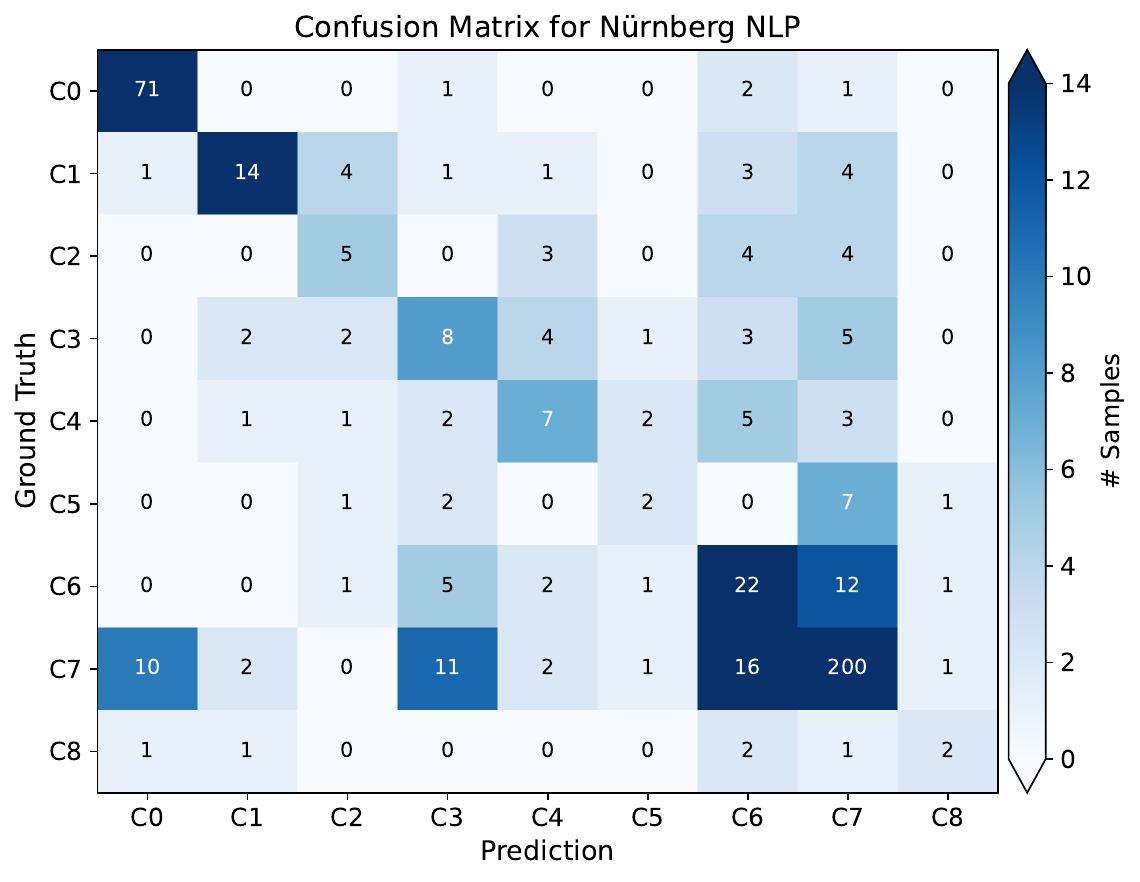}
        \caption{N\"urnberg NLP}
        \label{fig:heatmap1}
    \end{subfigure}
    \hfill
    \begin{subfigure}[b]{0.49\linewidth}
        \includegraphics[width=\textwidth]{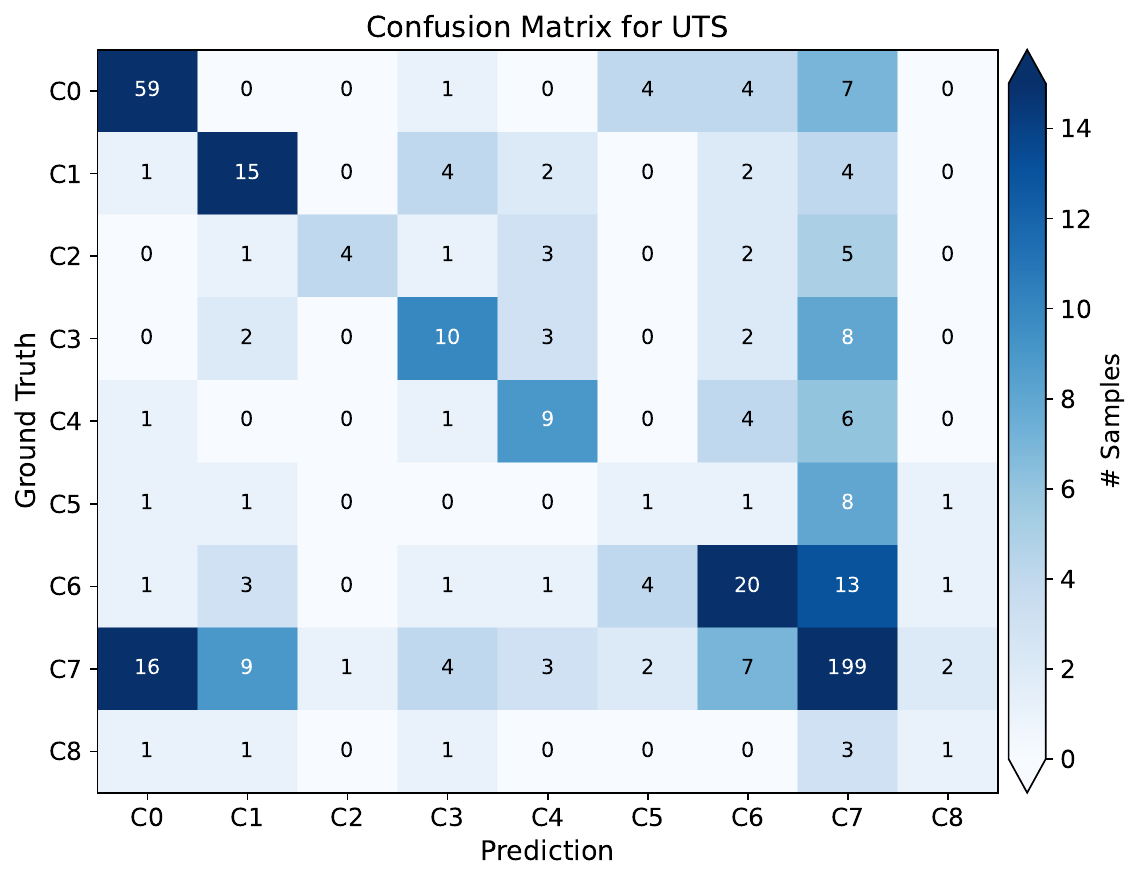}
        \caption{UTS}
        \label{fig:heatmap2}
    \end{subfigure}
    \begin{subfigure}[b]{0.49\linewidth}
        \includegraphics[width=\textwidth]{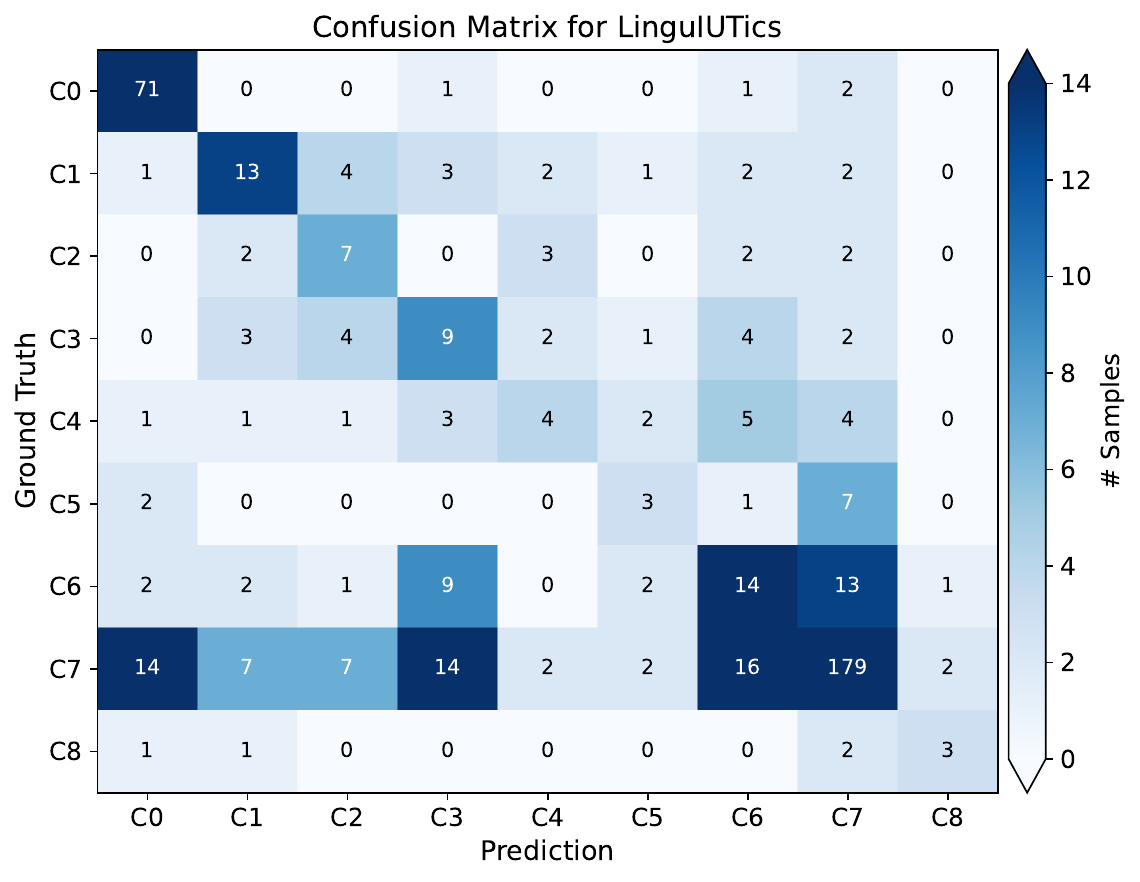}
        \caption{LinguIUTics}
        \label{fig:heatmap3}
    \end{subfigure}
    \hfill
    \begin{subfigure}[b]{0.49\linewidth}
        \includegraphics[width=\textwidth]{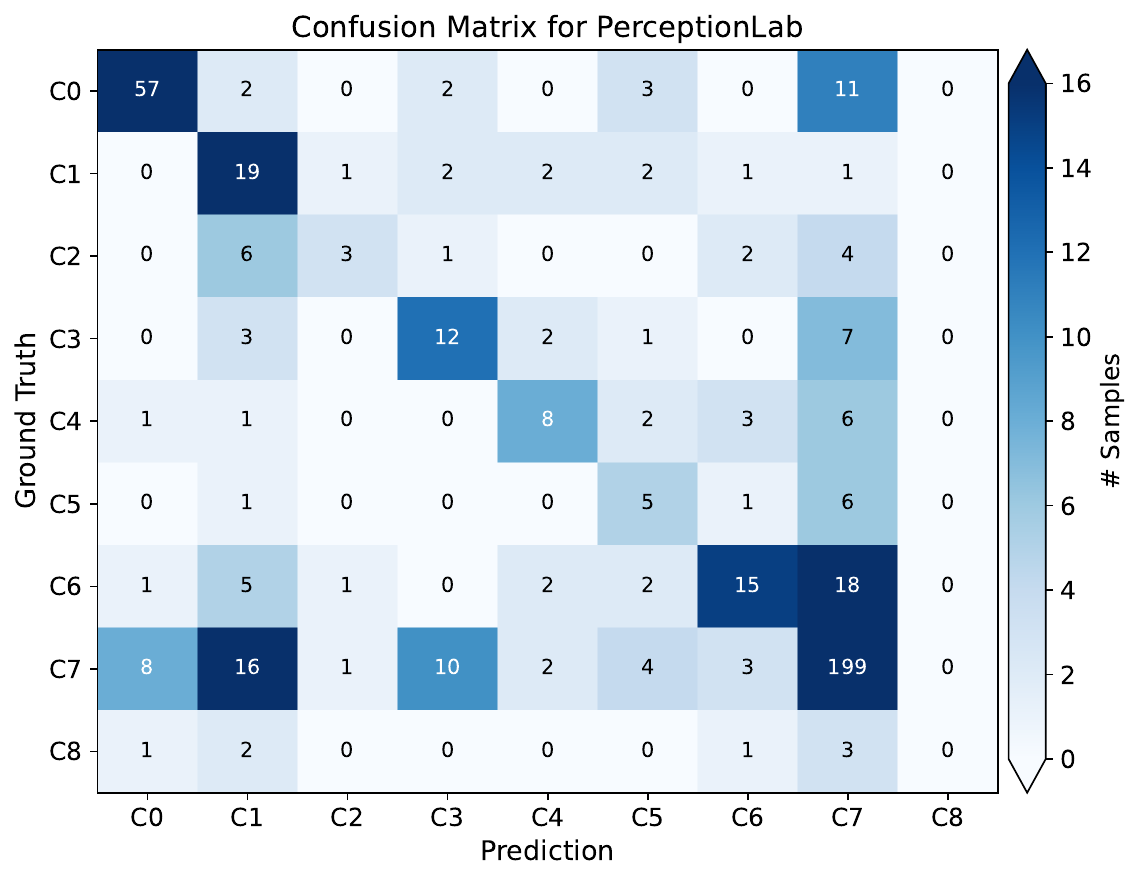}
        \caption{PerceptionLab}
        \label{fig:heatmap4}
    \end{subfigure}
    \caption{Confusion matrices for the four top-ranked systems on the
    held-out test set. Rows denote ground-truth labels, columns denote
    predictions. Color scales are clipped for readability; cell annotations
    show raw counts.}
    \label{fig:heatmap}
\end{figure*}
 
Figure~\ref{fig:heatmap} shows confusion matrices for the four
top-ranked systems. A single error pattern dominates all four:
systematic \emph{over-prediction of the High-Adaptive level} (C7),
the ``L7 attractor'' identified by the dataset paper as the central
failure mode of language models on \textsc{PsyDefConv}
\cite{na-etal-2026-psydefconv}. More interesting is \emph{how} the
four systems trade this attractor against minority-class recall.
\textsc{LinguIUTics} is the most conservative on C7 (diagonal of 179
vs.\ ${\approx}200$ for the others), a direct effect of its
minority-class augmentation and logit-bias calibration; the cost is
some C7 recall, but the benefit is improved recovery of Level~8 instances.
\textsc{PerceptionLab} shows a different signature: its dynamic
DMRS-Q retrieval produces markedly higher Level~1 recall (19/28),
while also exhibiting a wider over-prediction of C7. No single design
choice resolves the attractor; different strategies trade one form
of error against another, and the choice determines \emph{which}
minority classes a system can recover.
 
\subsection{What Worked, What Did Not}
 
Three takeaways emerge across the 15 system papers. First,
\emph{ensembling is the strongest differentiator at the top}: the
two highest-scoring systems both engineer error independence
explicitly---\textsc{Nürnberg NLP} along three orthogonal axes, and
\textsc{UTS} through a deliberative agent council---while
\textsc{TONI-NLP} \cite{paul-etal-2026-toninlp} reaches the same
conclusion through portfolio comparison. Second, \emph{task-specific
supervision beats backbone scale}: mid-scale backbones (Qwen3-8B,
Ministral-8B, Qwen3-14B) dominate the top half of the leaderboard,
and \textsc{AlienAnnotators}
\cite{karip-hossain-2026-alienannotators} report directly that scale
alone does not reliably help zero-shot performance whereas
clinically grounded prompts consistently do. Third, \emph{zero-shot
prompting alone is insufficient}: no system that relied primarily on
zero-shot prompting of a large LLM exceeded F1~=~0.30, and every described top-10
system combines prompting with fine-tuning, retrieval, or
ensembling. Clinical theory, packed into a prompt, is necessary but
must be coupled with task-specific supervision to translate into
accurate classification.
\section{Future Directions}
\label{sec:future}
Four directions stand out as natural next steps. (1)~\emph{Dialogue-level
defensive trajectories}: moving from utterance-level classification to
modeling how defenses evolve across a conversation, building on the
emotion-level trajectory analysis already included in the dataset
paper \cite{na-etal-2026-psydefconv}. (2)~\emph{Multilingual
\textsc{PsyDefConv}}: annotating comparable corpora in Chinese,
Japanese, Spanish, and other languages, with \textsc{DMRS Co-Pilot}
lowering the clinical-annotation cost that has historically blocked
cross-lingual scaling. (3)~\emph{Coupling defenses and supporter
strategies}: jointly modeling how supporters adapt their strategies
to seeker defensive functioning, closing the loop back to the broader
ESC research programme. (4)~\emph{Theory-aware generation}: using
defense-level predictions to condition emotionally supportive
responses, operationalizing the seeker-model $\rightarrow$
supporter-response direction outlined as future work in the dataset
paper. The first and fourth directions, in particular, turn
\textsc{PsyDefDetect} from a static classification benchmark into a
building block for clinically informed dialogue systems.
 
\section{Conclusion}
\label{sec:conclusion}
We introduced \textsc{PsyDefDetect}, the first shared task on
detecting DMRS-grounded psychological defense levels in emotional
support dialogues. The task drew 172 CodaBench participants, 21
ranked teams, and 15 system description papers across 12 countries.
The winning system reached macro-F1~=~0.420, a ${\approx}10.5$-point gain
over the strongest fine-tuned baseline, yet leaves clear headroom on
minority classes where the ``L7 attractor'' dominates. We release all
task materials to support continued work at the intersection of
clinical psychology and NLP.

\section*{Limitations}

This shared task has several limitations. First, \textsc{PsyDefConv} is a
relatively small corpus derived from English ESConv dialogues, so results may
not generalize to other languages, cultures, clinical settings, or naturally
occurring therapy conversations. Second, the task labels defenses at the DMRS
level rather than at the individual-mechanism level. This choice improves
annotation reliability, but it also collapses clinically meaningful distinctions
within each level. Third, defense interpretation remains inherently contextual:
even with trained annotators and adjudication, short seeker utterances can be
ambiguous without broader personal or longitudinal information. Finally, our
methodological analysis is based on the 15 teams that submitted system
description papers; leaderboard entries without accompanying papers are included
in the quantitative ranking but cannot be analyzed in the same level of detail.

\section*{Acknowledgments}
We thank all 21 participating teams and the 15 teams that contributed system
description papers for making this shared task a success. We thank the
BioNLP workshop organizers for hosting \textsc{PsyDefDetect}. We gratefully
acknowledge the \emph{Label Studio Academic Program} for providing access
to the annotation platform used to construct \textsc{PsyDefConv}. 
 
\section*{Ethical Considerations}
 
\paragraph{Data.}
\textsc{PsyDefConv} is derived from ESConv \cite{liu-etal-2021-towards} under
its stated terms. Only derived annotations, dialogue identifiers, and code
are redistributed; downstream users must obtain ESConv separately and
comply with its license.
 
\paragraph{Intended use.}
The dataset and resulting models are intended \emph{solely for research} on
language and defensive functioning. They are \textbf{not diagnostic tools}
and must not be used to make clinical, legal, or employment decisions about
individuals. The corpus is in English and reflects the domains covered by
ESConv, so results may not generalize across cultures or clinical settings.

\bibliography{custom}

\appendix



\end{document}